\documentclass{article}
\usepackage{spconf,amsmath,graphicx,hyperref}
\usepackage{amsmath, amssymb,adjustbox, booktabs, multirow}

\makeatletter
\renewcommand{\section}{\@startsection{section}{1}{0pt}
	{-2ex plus -.5ex minus -.2ex}
	{2ex plus .2ex}
	{\normalfont\large\bfseries}}
\makeatother

\makeatletter
\renewcommand{\subsection}{\@startsection{subsection}{2}{0pt}
	{-1.5ex plus -.4ex minus -.2ex}
	{1ex plus .2ex}%
	{\normalfont\normalsize\bfseries}}
\makeatother
\makeatletter
\renewcommand{\subsubsection}{\@startsection{subsubsection}{3}{0pt}
	{-1ex plus -.2ex minus -.1ex}
	{0.5ex plus .1ex}
	{\normalfont\normalsize}} 
\makeatother

\title{CONJUGATE RELATION MODELING FOR FEW-SHOT KNOWLEDGE GRAPH COMPLETION}

\name{Zilong Wang$^{1}$, Qingtian Zeng$^{1 \star}$, Hua Duan$^{1 \star}$, Cheng Cheng$^{1 \star}$, Minghao Zou$^{2}$, Ziyang Wang$^{3}$}
\address{$^{1}$Shandong University of Science and Technology, Qingdao, China\\$^{2}$ Cardiff University, Cardiff, United Kingdom\\$^{3}$Aston University, Birmingham, United Kingdom}
\begin{document}
	\maketitle
	\begin{abstract}
		Few-shot Knowledge Graph Completion (FKGC) infers missing triples from limited support samples, tackling long-tail distribution challenges. Existing methods, however, struggle to capture complex relational patterns and mitigate data sparsity. To address these challenges, we propose a novel FKGC framework for conjugate relation modeling (CR-FKGC). Specifically, it employs a neighborhood aggregation encoder to integrate higher-order neighbor information, a conjugate relation learner combining an implicit conditional diffusion relation module with a stable relation module to capture stable semantics and uncertainty offsets, and a manifold conjugate decoder for efficient evaluation and inference of missing triples in manifold space. Experiments on three benchmarks demonstrate that our method achieves superior performance over state-of-the-art methods.
	\end{abstract}
	\section{Introduction}
	\label{sec:intro}
	Knowledge Graph Completion (KGC) aims to infer missing entities or relations from existing graphs. Traditional knowledge graph embedding methods\cite{bordes2013translating, YangYHGD14a, trouillon2016complex} map entities and relations into low-dimensional spaces and have gained attention for their efficiency, yet they depend on abundant contextual information. Real-world graphs, however, often follow a long-tail distribution, where only a few entities or relations dominate connections\cite{xiong2018one}. To address this challenge, Few-shot Knowledge Graph Completion (FKGC) has been introduced.
	
	FKGC aims to learn relation or entity representations when the support set contains only a few triples and generalize them to the query set to predict missing information. Existing methods are generally categorized into two types: metric-based methods\cite{xiong2018one, zhang2020few, sheng2020adaptive} and meta-learning methods\cite{chen2019meta, niu2021relational, qiao2023relation}. Metric-based methods utilize neighborhood aggregation strategies to embed entities and compute distances between support and query set embeddings using predefined metric functions. Meta-learning methods integrate FKGC with meta-learning, enabling rapid adaptation to relational reasoning tasks by leveraging transferred knowledge. Although these have made some progress, they still have limitations in capturing complex relational patterns and handling sparse data\cite{luo2023normalizing}.

	Most methods model support triples as single deterministic embeddings, limiting their ability to capture the diversity of complex or incomplete relations. To address this limitation, some studies adopt distributional modeling. For example, UFKGC\cite{li2024uncertainty} represents entity and relation embeddings as Gaussian distributions and incorporates neighbor information via uncertainty-aware graph neural networks. NP-FKGC\cite{luo2023normalizing} and NFGN\cite{yuan2025normalizing} combine Gaussian embeddings with normalizing flows, replacing traditional meta-learning frameworks to capture complex distributions and uncertainty. However, they primarily rely on unconditional random divergence, which may produce overly diffuse latent distributions inconsistent with support sets, raising the risk of misclassification in few-shot scenarios.
	\begin{figure*}[htbp]
		\vspace{-1.5 em}
		\centering
		\includegraphics[width=0.85\textwidth]{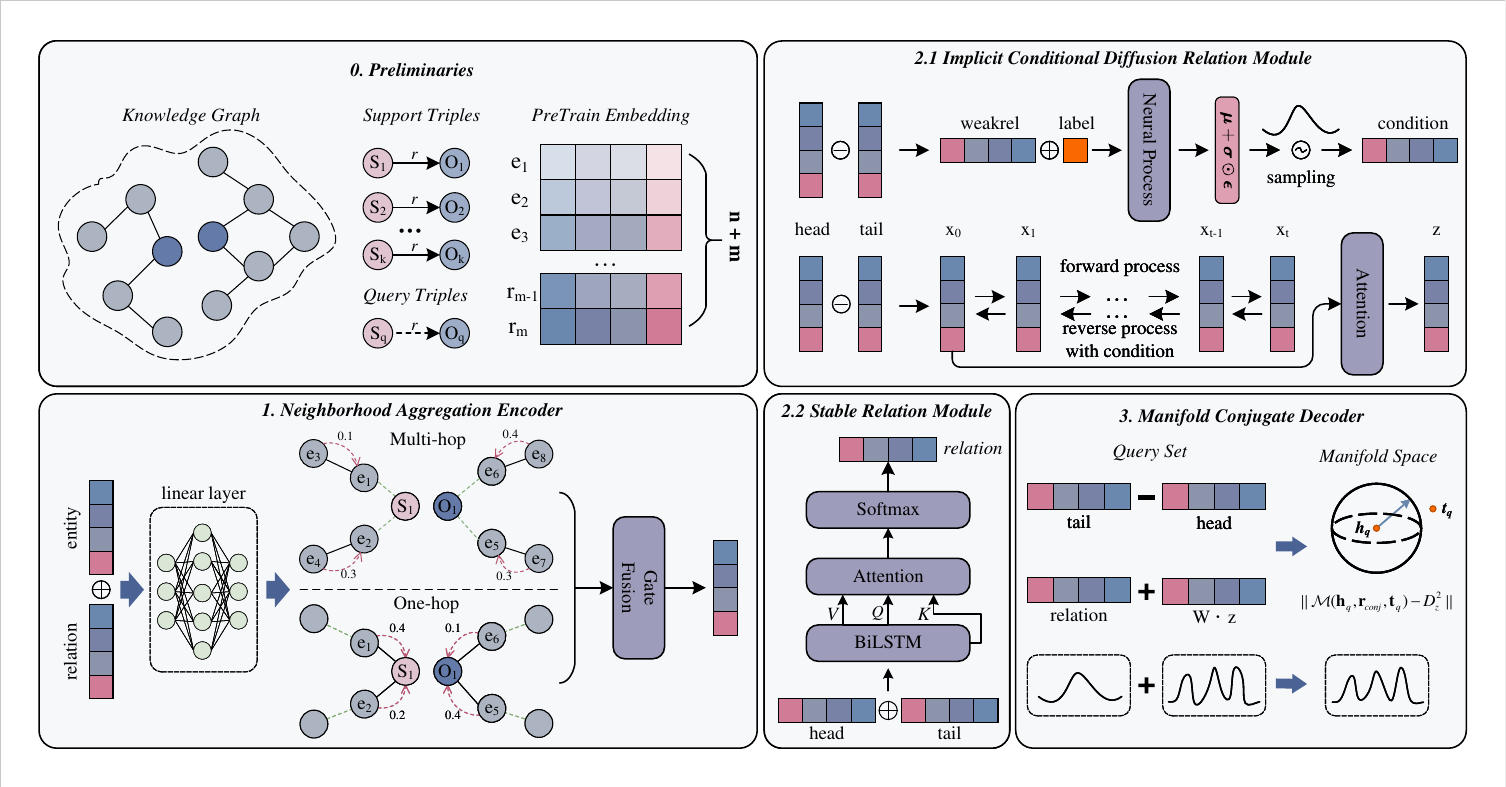}
		\vspace{-0.7 em}
		\caption{The overall framework of CR-FKGC. It consists of three main components: a neighbor aggregation encoder, a conjugate relation learner, and a manifold conjugate decoder.}
		\label{fig:model}
		\vspace{-1.3 em}
	\end{figure*}
	
	To address these challenges, we introduce a conjugate relation modeling framework for few-shot knowledge graph completion (CR-FKGC). Specifically, we first introduce a neighborhood aggregation encoder combining Relational Graph Attention Networks (RGAT)\cite{wang2025dynamic} with a gating mechanism to integrate high-order neighbor information effectively. We then design a conjugate relation learner comprising an implicit conditional diffusion relation module and a stable relation module. The former utilizes a score-based diffusion model to capture uncertainty offsets and employs neural processes\cite{garnelo2018conditional} to model the support set and labels, generating implicit conditional signals to guide the diffusion process. The latter employs BiLSTM with a attention mechanism to derive a stable relation representation. Finally, a manifold conjugate decoder is designed to jointly model stable semantics and uncertainty offsets in manifold space, enabling efficient evaluation and reasoning of missing triples.
	
	Contributions: (1) We introduce CR-FKGC, a conjugate relation modeling framework for FKGC. To our knowledge, we are the first to introduce conjugate relations and employ conditional guidance for uncertainty modeling. (2) We design a conjugate relation learner that jointly models stable semantics and uncertainty offsets for relations. (3) We propose a manifold conjugate decoder that captures complex relational semantics in manifold space. (4) We propose a neighborhood aggregation encoder that combines RGAT with a gating mechanism to integrate high-order neighbor information while suppressing redundancy and noise. (5) Extensive experiments on three public datasets demonstrate that CR-FKGC consistently outperforms state-of-the-art baselines.
	\section{method}
	\label{sec:method}
	Given a knowledge graph (KG) $\mathcal{G} = \{(h, r, t) \mid h, t \in \mathcal{E}, r \in \mathcal{R}\}$, each relation $r$ is associated with a support set $\mathcal{S} = \{(h_i, r, t_i)\}_{i=1}^{K}$ and a query set $\mathcal{Q} = \{(h_j, r, ?)\}$. The objective is to predict the missing tail entity $t_j$ in the query set based on $\mathcal{S}$. To address this task, we propose CR-FKGC, a framework for conjugate relation modeling in few-shot knowledge graph completion. As illustrated in Fig.~\ref{fig:model}, it comprises three core components: the Neighborhood Aggregation Encoder, the Conjugate Relation Learner (CRL), and the Manifold Conjugate Decoder (MaConE).
	\subsection{Neighborhood Aggregation Encoder}
	To expand the receptive field of the target node and suppress distant neighbor noise, we propose a neighborhood aggregation encoder that combines a gating mechanism with RGAT. For the neighbor set $\mathcal{N}_h$ of target node $h$, each neighbor entity $\mathbf{e}_i$ is concatenated with its relation embedding $\mathbf{r}_i$ and mapped into a neighborhood vector $\mathbf{m}_i = W_r[\mathbf{e}_i;\mathbf{r}_i] + b_r$, where $W_r$ and $b_r$ are trainable parameters. The aggregated neighborhood representation $\mathbf{h}_{\text{a}}$ is then computed through attention:
	\begin{gather}
		\alpha_i = \frac{\exp({\text{LeakyReLU}(\mathbf{W}^{\top}} [\mathbf{h};\mathbf{m}_i]))}{\sum_{j \in \mathcal{N}_h}\exp({\text{LeakyReLU}(\mathbf{W}^{\top}} [\mathbf{h};\mathbf{m}_j]))},\\
		\mathbf{h}_{\text{a}} = \sum_{i \in \mathcal{N}_h} \alpha_i \cdot \mathbf{m}_i,
	\end{gather}
	where $\mathbf{W}$ is a trainable weight matrix and $[\cdot;\cdot]$ denotes vector concatenation. Finally, gated fusion adaptively combines neighborhood and original features:
	\begin{gather}
		\textbf{h}^{'} = \text{ReLU}\big(\textbf{g} \odot \textbf{h}_{\text{a}} + (1 - \textbf{g}\big) \odot \textbf{h}),\quad \mathbf{g} = \sigma(\textbf{W}_g\mathbf{h}_{\text{a}} + b_g),
	\end{gather}
	where $\sigma(\cdot)$ is the sigmoid function, $\odot$ denotes element-wise multiplication, and $\textbf{W}_g$ and $b_g$ are trainable parameters.
	\subsection{Conjugate Relation Learner}
	The coexistence of stable semantics and uncertainty poses a critical challenge in FKGC. Therefore, we introduce the concept of conjugate relations, where relations are represented as a pair of a stable relation representation and an uncertainty offset. Building on this, we design a conjugate relation learner consisting of the Implicit Conditional Diffusion Relation (ICDR) module and the Stable Relation (SR) module.
	\subsubsection{Implicit Conditional Diffusion Relation Module}
	As shown in Fig.~\ref{fig:model}, we design an implicit conditional diffusion relation module to model the uncertainty of latent patterns. The module consists of two stages: Uncertainty Diffusion Modeling and Implicit Conditional Generation.
	
	\noindent \textbf{Uncertainty Diffusion Modeling}.~The uncertainty modeling process is formulated as a continuous-time diffusion model based on Stochastic Differential Equation (SDE)\cite{song2020score, lu2022dpm}.
	For each support triple $(h, r, t)$, we construct a weak relation feature $\textbf{h}_{\text{weak}} = \textbf{h}_t - \textbf{h}_h$ and concatenate them to form the initial relation embedding $\mathbf{x}_0$. During the forward diffusion process, $\mathbf{x}_0$ gradually evolves into Gaussian noise $\mathbf{x}_T$.
	This process can be modeled by the following Itô-form SDE:
	\begin{gather}
		d\textbf{x}_t = f(\textbf{x}_t, t)dt + g(t)d\textbf{w}_t,
	\end{gather}
	where $f(\cdot, t)$ is the drift term, $g(t)$ is the time-dependent diffusion coefficient, and $\textbf{w}_t$ is the standard Brownian motion.
	During the reverse process, starting from $\mathbf{x}_T$, $\mathbf{x}_0$ is gradually restored under the guidance of implicit conditional $\mathbf{c}$. Its reverse SDE is expressed as:
	\begin{gather}
		d\textbf{x}_t = [f(\textbf{x}_t, t) - g(t)^2 \nabla_{\textbf{x}_t} \log p_t(\textbf{x}_t \mid \textbf{c})]dt + g(t) d\bar{\textbf{w}}_t,
	\end{gather}
	where  $p_t(\textbf{x}_t | \textbf{c})$ is the intermediate state distribution conditioned on $\textbf{c}$, the score function $\nabla{\mathbf{x}_t} \log p_t(\cdot)$ is approximated by a neural network $\epsilon_\theta(\textbf{x}_t, t, \textbf{c})$ based on a U-Net architecture, and $x_0$ is then processed by the attention pooling to yield the uncertainty offset $z$.
	
	\noindent \textbf{Implicit Conditional Generation}.~Instead of explicitly concatenating support set information as a control condition, we model the joint distribution of support triplets via a neural process to generate a relation-specific implicit conditional vector. Specifically, we first concatenate the weak relation features with the corresponding labels $y_i \in \{0,1\}$ and input them into Neural Process Encoder (NPEncoder) to obtain the latent representation of the triple ${s_i}$. We then perform average pooling on all ${s_i}$ to obtain the global context variable of relation $\textbf{r}$:
	\begin{gather}
		\mathbf{r} = \frac{1}{|\tilde{S}_r|} \sum_{\textbf{s}_i \in \tilde{S}_r} \mathbf{s}_i, \quad \textbf{s}_i = \text{NPEncoder}([\textbf{h}_{weak}^i; y_i]),
	\end{gather}
	where the NPEncoder is implemented as an MLP. Next, we map $\mathbf{r}$ to a Gaussian latent distribution, predict the mean $\boldsymbol{\mu}$ and standard deviation $\boldsymbol{\sigma}$ through two MLPs, and employ the reparameterization trick to sample the implicit conditional vector $\mathbf{c}$:
	\begin{gather}
		\boldsymbol{\mu} = \text{MLP}(\mathbf{r}), \quad \boldsymbol{\sigma} = 0.1 + 0.9 \cdot \sigma(\text{MLP}(\mathbf{r})), \\
		\mathbf{c}= \boldsymbol{\mu} + \boldsymbol{\sigma} \odot \boldsymbol{\epsilon}, \quad \boldsymbol{\epsilon} \sim \mathcal{N}(\mathbf{0}, \mathbf{I}),
	\end{gather}
	where $\sigma(\cdot)$ is the sigmoid activation function, $\boldsymbol{\epsilon}$ is random noise sampled from a standard normal distribution.
	
	To optimize the implicit conditional diffusion relation module, we define a joint loss comprising two components: a diffusion denoising loss, which measures the discrepancy between the predicted noise $\epsilon_\theta(x_t, t, c)$ and the ground-truth noise $\epsilon$, and a KL divergence term that regularizes the posterior distribution $p(s \mid S_r)$ inferred from the support set to be closer to the actual distribution $q(s)$:
	\begin{gather}
		\mathcal{L}_{rel} = \mathbb{E}_{x_0, \epsilon, t} \left[ \|\epsilon - \epsilon_\theta(\textbf{x}_t, t, c)\|^2 \right] + D_{\text{KL}}( q(s) \,\|\, p(s \mid S_r) ),
	\end{gather}
	where $S_r$ and $Q_r$ denote the support set and query set of relation $r$, respectively. As $q(s)$ is generally intractable, we approximate it using $p(s \mid S_r, Q_r)$.
	\subsubsection{Stable Relation Module}
	To learn semantically stable relation representation, we employ a BiLSTM with an attention mechanism as a stable relation module. Specifically, we concatenate the head and tail entity embedding to obtain a preliminary relation sequence $[\mathbf{r}_s^1, \dots, \mathbf{r}_s^K]$, which is then encoded by the BiLSTM:
	\begin{gather}
		\textbf{r}_K, \dots , \textbf{r}_1 = \text{BiLSTM}(\textbf{r}_s^K, \dots, \textbf{r}_s^1),
	\end{gather}
	Subsequently, we employ a attention pooling  and a linear layer to obtain the stable relation embedding:
	\begin{gather}
		\textbf{r}_s = \text{FC}(\hat{\textbf{r}}), \quad \hat{\textbf{r}} = \sum_{i=1}^{K} \frac{\exp(\tanh(\textbf{W}_s r_i + b_s))}{\sum_{j=1}^{K} \exp(\tanh(\textbf{W}_s r_j + b_s))} \, r_i,
	\end{gather}
	where $\textbf{W}_s$ and $b_s$  are the trainable weight matrix and bias term, respectively.
	\subsection{Manifold Conjugate Decoder}
	Existing knowledge graph decoders, such as TransE or cosine similarity-based decoders, rely on deterministic relation embeddings, which are inadequate for the conjugate relation structure. Therefore, we propose a Manifold Conjugate Decoder (MaConE) that operates in manifold space. Each relation $r$ is represented as a conjugate pair: a stable relation representation $\mathbf{r}_s$, and an uncertainty offset $\mathbf{z}$. The offset is projected via a fully connected layer and added to the stable representation $\textbf{r}_{conj} = \textbf{r}_s + \text{FC}(\textbf{z})$, where $\mathrm{FC}(\cdot)$ is a fully connected layer. The scoring function maps a triple $(h_q, r, t_q)$ into a manifold space for semantic evaluation:
	\begin{gather}
		\text{score}(h, r, t) = \Vert \mathcal{M}(\textbf{h}_q, \textbf{r}_{conj}, \textbf{t}_q)  - D^2_z \Vert,
	\end{gather}
	where $D_z=\text{MLP}(\mathbf{c})$ generates a dynamic boundary threshold from the implicit condition $\mathbf{c}$, The manifold function $\mathcal{M}(\cdot)$ is instantiated with TransE.
	
	The decoder is optimized using a margin-based ranking loss. Given positive triples $\mathcal{Q}_r^+$ and negative triples $\mathcal{Q}_r^-$ for relation $r$, the function is defined as:
	\begin{align}
		\mathcal{L}_{tri} & = \sum_{r} \sum_{(h_q, t_q^+) \in \mathcal{Q}_r^+}
		\sum_{(h_q, t_q^-) \in \mathcal{Q}_r^-} \notag                                                                  \\
		& \quad \max\Big(\gamma + \text{score}(h_q, r, t_q^-) - \text{score}(h_q, r, t_q^+), 0 \Big),
	\end{align}
	where $\gamma$ is the margin between positive and negative samples, typically set to 1. The overall training objective is formulated as $\mathcal{L} = \mathcal{L}_{\text{tri}} + \mathcal{L}_{\text{rel}}$.
	\section{EXPERIMENTS}
	\label{sec:experiments}
	\subsection{Experimental Setting}
	\textbf{Datasets and Evaluation Metrics}. Following prior work, we evaluate the proposed method on three benchmark datasets: NELL-One\cite{mitchell2018never}, FB15k237-One\cite{toutanova2015representing}, and Wiki-One\cite{vrandevcic2014wikidata}, and the Hits@N and MRR are metrics used for evaluation.
	
	\begin{table*}[t]
		\vspace{-2 em}
		\centering
		\caption{Overall Performance of 5-Shot Link Prediction on NELL-One, FB15K237-One, and Wiki-One.}
		\begin{adjustbox}{width=1\textwidth}
			\renewcommand{\arraystretch}{0.9}
			\normalsize
			\begin{tabular}{lcccccccccccc}
				\toprule
				\multirow{2}[2]{*}{Model} & \multicolumn{4}{c}{NELL-One} & \multicolumn{4}{c}{FB15K237-One} & \multicolumn{4}{c}{Wiki-one}                                                                                                                                                                                     \\
				\cmidrule{2-13}           & MRR                          & Hit @ 10                      & Hit @ 5                      & Hit @ 1           & MRR               & Hit @ 10          & Hit @ 5           & Hit @ 1           & MRR               & Hit @ 10          & Hit @ 5           & Hit @ 1           \\
				\midrule
				TransE(2013)             & 0.168                        & 0.345                         & 0.186                        & 0.082             & 0.307             & 0.537             & 0.419             & 0.198             & 0.052             & 0.090             & 0.057             & 0.042             \\
				DistMult(2015)             & 0.214                        & 0.319                         & 0.246                        & 0.140             & 0.237             & 0.378             & 0.287             & 0.164             & 0.077             & 0.134             & 0.078             & 0.035             \\
				ComplEx(2016)     & 0.239                        & 0.364                         & 0.253                        & 0.176             & 0.238             & 0.370             & 0.281             & 0.169             & 0.070             & 0.124             & 0.063             & 0.030             \\
				\midrule
				GMatching(2018)           & 0.176                        & 0.294                         & 0.233                        & 0.113             & 0.304             & 0.456             & 0.410             & 0.221             & 0.263             & 0.387             & 0.337             & 0.197             \\
				MetaR(2019)      & 0.261                        & 0.437                         & 0.350                        & 0.168             & 0.403             & 0.647             & 0.551             & 0.279             & 0.221             & 0.302             & 0.264             & 0.178             \\
				GANA(2021)                & 0.344                        & 0.517                         & 0.437                        & 0.246             & 0.458             & 0.656             & 0.575             & 0.349             & 0.351             & 0.446             & 0.407             & 0.299             \\
				RANA(2023)                & 0.361                        & \underline{0.573}             & 0.475                        & 0.253             & -                 & -                 & -                 & -                 & 0.379             & 0.480             & 0.437             & 0.329             \\
				NP-FKGC(2023)             & \underline{0.460}            & 0.494                         & 0.471                        & \underline{0.437} & \underline{0.538} & \underline{0.671} & \underline{0.593} & \underline{0.476} & \underline{0.503} & \underline{0.668} & \underline{0.599} & 0.423             \\
				UFKGC(2024)               & 0.417                        & \textbf{0.588}                & 0.511                        & 0.324             & -                 & -                 & -                 & -                 & 0.431             & 0.540             & 0.491             & 0.375             \\
				PARE(2025)                & 0.451                        & 0.503                         & 0.458                        & 0.365             & 0.502             & 0.664             & 0.589             & 0.428             & 0.478             & 0.553             & 0.583             & \underline{0.429} \\
				StarRing(2025)            & 0.421                        & 0.568                         & \underline{0.512}            & 0.340             & 0.396             & 0.598             & 0.518             & 0.300             & 0.406             & 0.551             & 0.488             & 0.330             \\
				\midrule
				CR-FKGC                   & \textbf{0.534}               & 0.568                         & \textbf{0.540}               & \textbf{0.516}    & \textbf{0.595}    & \textbf{0.752}    & \textbf{0.696}    & \textbf{0.508}
				& \textbf{0.561}               & \textbf{ 0.725}               & \textbf{ 0.638}              & \textbf{0.484}                                                                                                                                                                    \\
				\bottomrule
			\end{tabular}%
		\end{adjustbox}
		\label{tab:main}%
		\vspace{-1 em}
	\end{table*}%
	
	\noindent \textbf{Baseline Models}. We compare CR-FKGC with the following FKGC methods: TransE\cite{bordes2013translating}, DistMult\cite{YangYHGD14a}, ComplEx\cite{trouillon2016complex}, GMatching\cite{xiong2018one}, MetaR\cite{chen2019meta}, GANA\cite{niu2021relational}, RANA\cite{qiao2023relation}, PARE\cite{yu2025path}, NP-FKGC\cite{luo2023normalizing}, UFKGC\cite{li2024uncertainty}, and StarRing\cite{zhao2025few}.

	\subsection{Main Results and Analysis}
	Table \ref{tab:main} shows the performance comparison between CR-FKGC and baseline methods. For fair comparison, all results are reported from the original papers.  Specifically, CR-FKGC outperforms state-of-the-art methods across all three datasets, achieving gains of 7.4\%, 5.7\%, and 5.8\% in MRR and 7.9\%, 3.2\%, and 6.1\% in Hits@1, while also attaining superior performance on Hits@5 and Hits@10. This demonstrates that the proposed model not only outperforms the overall ranking performance but also has stronger discriminative power in tail entity recognition accuracy. This performance improvement is primarily due to the synergistic effect of the conjugate relation learner and the neighborhood aggregation encoder: the former effectively alleviates the semantic ambiguity of limited samples by simultaneously modeling relational stable semantics and uncertainty offsets; the latter expands the entity receptive field and suppresses redundant noise, improving semantic integrity. In addition, MaConE maps triples into a manifold space, further enhancing the ability to model complex relations. Notably, CR-FKGC achieves more significant improvements than uncertainty modeling methods such as NP-FKGC and UFKGC. This is due to both the uncertainty modeling advantage of the diffusion model in the denoising process and the implicit conditional embedding generated by the neural process, which makes the model more adaptable under limited samples.
	\begin{table}[h]
		\vspace{-1 em}
		\centering
		\caption{Ablation Study of Different Components.}
		\renewcommand{\arraystretch}{0.8}
		\begin{adjustbox}{width=0.48\textwidth}
			\normalsize
			\begin{tabular}{lcccccc}
				\toprule
				\multirow{2}[2]{*}{Variants} & \multicolumn{3}{c}{NELL-One} & \multicolumn{3}{c}{FB15K237-One}                                                                     \\
				\cmidrule{2-7}
				& MRR                          & Hit@10                        & Hit@1          & MRR            & Hit@10         & Hit@1          \\
				\midrule
				w/o Gate                     & 0.512                        & 0.564                         & 0.484          & 0.559          & 0.743          & 0.466          \\
				w/o Condition                & 0.513                        & 0.550                         & 0.485          & 0.548          & 0.737          & 0.468          \\
				w/o ICDR                     & 0.504                        & 0.547                         & 0.475          & 0.536          & 0.679          & 0.456          \\
				w/o SR                       & 0.470                        & 0.498                         & 0.453          & 0.510          & 0.674          & 0.436          \\
				w/o MaConE                   & 0.528                        & 0.554                         & 0.510          & 0.586          & 0.732          & 0.494          \\
				\midrule
				CR-FKGC                      & \textbf{0.534}               & \textbf{0.568}                & \textbf{0.516} & \textbf{0.595} & \textbf{0.752} & \textbf{0.508} \\
				\bottomrule
			\end{tabular}%
		\end{adjustbox}
		\label{tab:ablation}%
	\end{table}
	\subsection{Ablation Study and Parameter Analysis}
	\noindent \textbf{Ablation Study}. Table \ref{tab:ablation} presents ablation experiment results. Removing the gating mechanism (w/o Gate), performance declines due to increased noise and weaker entity representations. Replacing the implicit conditional embedding with a simple concatenation of the support set and its labels (w/o Condition) leads to a significant performance drop, confirming the necessity of neural process for conditional guidance. Removing either uncertainty offsets (w/o ICDR) or stable representations (w/o SR) degrades performance, indicating that both are indispensable for relation modeling. Replacing MaConE with TransE (w/o MaConE) causes a moderate decline in performance, demonstrating the effectiveness of MaConE in modeling conjugate relation representations.
	
\begin{figure}[h]
		\vspace{-0.5 em}
		\centering
		\includegraphics[width=0.48\textwidth]{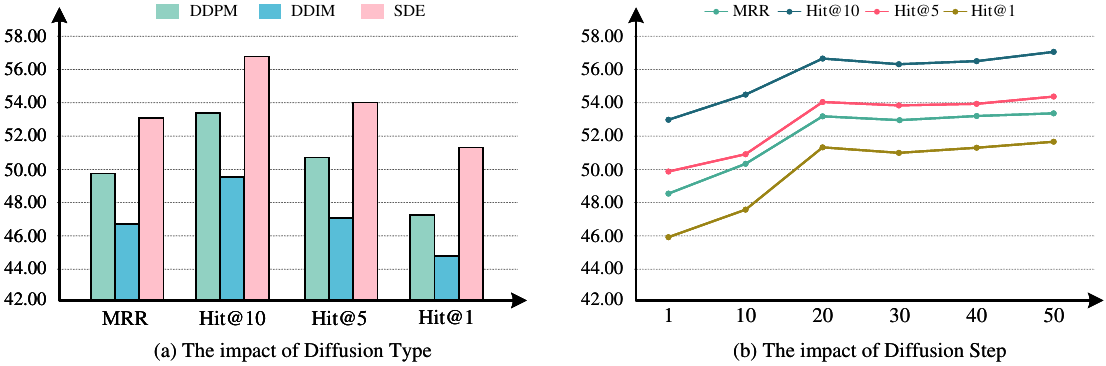}
		\caption{Impact of Diffusion Types and Steps on NELL-One.}
		\label{fig:diffusion}
		\vspace{-1 em}
	\end{figure}

	\noindent \textbf{Parameter Analysis}. We compared the performance impact of different diffusion model types and diffusion step numbers. As shown in the Fig.~\ref{fig:diffusion}, SDE is more capable of capturing subtle differences in data distribution than DDPM\cite{ho2020denoising} and DDIM\cite{song2020denoising}. Furthermore, an appropriate number of diffusion steps can improve the performance of relational uncertainty modeling. Considering both accuracy and computational overhead, we ultimately set the number of diffusion steps to 20.
	
	\section{CONCLUSION}
	\label{sec:conclusion}
	This paper proposes a novel FKGC method for conjugate relation modeling. The model designs a conjugate relation learner to jointly model relational stable semantics and uncertainty offsets. A manifold conjugate decoder is designed to integrate conjugate relation representations in manifold space to enable complex relation reasoning. Additionally, we integrate RGAT with a gating mechanism in neighborhood aggregation encoder to expand the receptive field of entities and constrain noise and redundant information. Experiments on three benchmarks validate the effectiveness of our method.
	
	\newpage
	\section{Acknowledgements}
	This work is supported by National Science and Technology Major Project of China [2022ZD0119501]; NSFC [52574256, 52374221, 62402261 and 72404177]; Sci. \& Tech. Development Fund of Shandong Province of China [ZR2022MF288, ZR2023MF097]; the Taishan Scholar Program of Shandong Province[TSTP20250506].
	{
		
		\fontsize{9pt}{11pt}\selectfont
		\bibliographystyle{IEEEbib}
		\bibliography{refs}
	}
	
\end{document}